# A Cross-Modal Rumor Detection Scheme via Contrastive Learning by Exploring Text–Image internal Correlations

## Authors

Bin Ma, Yifei Zhang, Yongjin Xian, Qi Li, Linna Zhou, Gongxun Miao

## Abstract

Existing rumor detection methods often neglect the content within images as well as the inherent relationships between contexts and images across different visual scales, thereby resulting in the loss of critical information pertinent to rumor identification. To address these issues, this paper presents a novel cross-modal rumor detection scheme based on contrastive learning, namely the Multi-scale Image and Context Correlation exploration algorithm (MICC). Specifically, we design an SCLIP encoder to generate unified semantic embeddings for text and multi-scale image patches through contrastive pretraining, enabling their relevance to be measured via dot-product similarity. Building upon this, a Cross-Modal Multi-Scale Alignment module is introduced to identify image regions most relevant to the textual semantics, guided by mutual information maximization and the information bottleneck principle, through a Top-K selection strategy based on a cross-modal relevance matrix constructed between the text and multi-scale image patches. Moreover, a scale-aware fusion network is designed to integrate the highly correlated multi-scale image features with global text features by assigning adaptive weights to image regions based on their semantic importance and cross-modal relevance. The proposed methodology has been extensively evaluated on two real-world datasets. The experimental results demonstrate that it achieves a substantial performance improvement over existing state-of-the-art approaches in rumor detection, highlighting its effectiveness and potential for practical applications.

## 1. Introduction

With the widespread use of social media, online rumors are increasingly taking the form of integrated textual and visual content, exhibiting strong deceptive characteristics that significantly compromise the credibility of information and erode public trust [1][2].

In recent years, multimodal rumor detection has leveraged diverse strategies to exploit complementary information across modalities, significantly boosting detection performance [3]. H et al. [4] introduced COMFUSE, a few-shot multimodal fusion model for detecting emerging

COVID-19 rumors with minimal labeled data. Goyal et al. [5] employed Graph Convolutional Networks (GCNs) by integrating textual, visual, and social network features to identify fake Twitter accounts. Li et al. [6] proposed FoRM, a two-phase framework combining coarse selection and fine-grained relational reasoning via attention mechanisms. While these approaches utilize multimodal cues, they often neglect cross-modal correlations. Yadav et al. [7] developed ETMA, which employs hierarchical attention to associate textual descriptions with key visual regions. MCAN, proposed by Qian et al. [8], incorporates co-attention layers to model the cognitive process of humans interpreting text-image news, reinforcing modality interdependence. However, existing models still struggle with intra-image variability. First, they cannot selectively extract text-relevant visual cues, overlooking semantically important image regions. Second, unrelated visual elements may act as noise, compromising detection accuracy. Xue et al. [9] focused on inter-modal consistency by computing cosine similarity across modalities, hypothesizing that high similarity implies truthfulness. However, this assumption may oversimplify the complex relationships between text and images, overlooking subtle but semantically relevant visual cues that diverge from textual content. These cues, though not strictly similar, are highly correlated and critical for accurate rumor detection, highlighting the need to model cross-modal relevance rather than mere similarity. Overall, current methods require refinement in modeling modality correlation, managing visual variance, and suppressing noise.

With the advancement of pretrained cross-modal models such as CLIP and Chinese CLIP [10][11], contrastive learning between images and text has emerged as an effective approach for constructing a unified semantic space. Zhou et al. [12] proposed a CLIP-guided multimodal learning framework that improves fake news detection by leveraging image–text semantic alignment. Although such models perform well in coarse-grained image-text retrieval tasks, they struggle to extract discriminative region-level visual features. ViT-based CLIP models can also only capture fine-grained features at a single scale; integrating external image segmentation modules may enhance precision but at the cost of significantly increased computational overhead. To effectively capture image region features at multiple granularities, inspired by GoogLeNet's use of parallel multi-scale convolutions to enhance model expressiveness by covering diverse receptive fields [13], we design the SCLIP encoder (Scale-aware Contrastive Language–Image Projection). This encoder integrates the semantic alignment strengths of contrastive learning with multi-scale convolution to extract localized visual semantics at different granularities, which are then aligned with global textual semantics in a cross-modal contrastive pretraining framework, enabling comprehensive modeling of visual information and consistent alignment in the shared semantic space.

Cross-modal semantic alignment is widely regarded as a crucial element for improving rumor detection performance. Peng et al. [14] validated its effectiveness in the MRML model by employing triplet learning to model intra-modal relationships and using contrastive learning to capture inter-modal associations. To further enhance alignment quality, our method is grounded in two principles: Mutual Information Maximization[15], which encourages the model to retain image regions that share the most semantic information with the text, ensuring cross-modal consistency; and the Information Bottleneck[16], which seeks compact and task-relevant representations by minimizing redundant visual information while preserving critical semantics. Based on the fine-grained visual features provided by the SCLIP encoder, we introduce the Cross-Modal Multi-scale Alignment module, which adopts dot-product as a relevance metric to construct a cross-modal relevance matrix. Guided by the above principles, the model then performs Top-K selection from multi-scale visual regions to identify K regions most semantically aligned with the text within each scale. This strategy effectively suppresses noise, compresses redundancy, and strengthens fine-grained semantic alignment across modalities.

When fusing the top-K highly relevant local image features with a single global textual representation, an imbalance in cross-modal weight allocation may occur. Direct fusion may result in the suppression of textual information during downstream processing, causing important features to be overlooked. To address this, we propose the Scale-Aware Fusion Network, which employs a feed-forward network to learn semantic importance scores and integrates them with the relevance scores from the Cross-Modal Multi-scale Alignment module. The resulting importance weights for different regions are then used to perform weighted fusion with the global textual feature. This design effectively mitigates the issue of improper cross-modal weighting and enhances the model's ability to integrate multimodal information.

The main contributions of this paper are as follows:

●We propose the SCLIP encoder, which enables multi-granular visual representation learning via multi-scale convolutional receptive fields, without relying on explicit image segmentation. Coupled with a Transformer for semantic encoding of text and multi-scale visual patches, the model is pretrained using contrastive learning to align modalities in a shared semantic space. This approach effectively balances representational detail and computational efficiency.

●We develop a Cross-Modal Multi-scale Alignment module that measures semantic correlation between modalities using dot-product relevance to construct a relevance matrix between image regions and text. Guided by mutual information maximization and the information bottleneck principle, the module selects the Top-K image regions most aligned with the text, effectively preserving critical semantic cues while suppressing redundant information.

●To capture the semantic and relevance differences across multi-scale image regions, we introduce the Scale-Aware Fusion Network. This module learns semantic-level scores via a feed-forward network and integrates them with the correlation scores from the relevance matrix to generate weighted representations for image regions. The refactored representations are then fused with textual features, thereby improving the expressiveness of cross-modal representations.

This paper is structured as follows. Section II surveys existing literature on rumor detection. Section III details the proposed framework. Section IV discusses the experimental settings and provides further analysis of parameter configurations, followed by comparative evaluations and ablation studies to demonstrate the efficacy of the proposed method. Section V concludes the work and discusses potential avenues for future investigation.

## 2. Related Work

In response to the growing complexity of rumor formats, various rumor detection techniques have been developed. Early approaches focused mainly on analyzing textual features to identify rumors through linguistic patterns. Recently, multimodal rumor detection has attracted increasing interest, as these methods combine information from text, images, and social context to improve detection accuracy [17]. At present, rumor detection heavily relies on extracting diverse features from multimodal social media content, including textual posts, visual data, social interactions, and associated knowledge graphs [18]. Therefore, this study categorizes and reviews related work in four areas: text-based analysis, text-image fusion, social network analysis, and external knowledge or fact-based approaches.

Text-based approaches primarily detect rumors by leveraging natural language processing techniques to uncover latent linguistic cues. Early works relied on handcrafted features such as bag-of-words and TF-IDF [19] combined with traditional classifiers like SVM and random forest. Ma et al. [20] introduced a rule-based method with linear classifiers, which, while effective, depended heavily on manual feature engineering and lacked the capacity to capture deep semantic structures. With the advent of deep learning, models including RNNs, LSTMs, and CNNs became prominent in rumor detection. For example, Ma et al. [21] used RNNs to model the temporal characteristics of Weibo posts, and Yu et al. [22] applied CNNs to identify salient textual patterns. Yao et al. [23] proposed TextGCN, a graph-based approach that encodes document-level structure to enhance semantic representation. More recently, pretrained language models like BERT [24] have gained popularity for their strong contextual learning, especially under limited data and transfer learning scenarios. Tian et al. [25] introduced the QSAN model, which incorporates a quantum symbolic attention mechanism, combining quantum encoding with semantic inference and expanding the theoretical scope of text modeling.

To better understand rumors in social media, researchers have explored joint modeling of textual and visual data, referred to as text-image based methods. The CARMN model [26] integrates multimodal features using a Cross-modal Attention Residual Network (CARN) and a Multi-Channel CNN (MCN), while att_RNN [27] enhances semantic fusion through attention-based cross-modal aggregation. Khattar et al. [28] proposed MVAE, a multimodal variational autoencoder that fuses textual and visual representations to improve fake news detection. To bridge modality gaps, Hua et al. [29] applied contrastive learning to reinforce semantic alignment between images and text. Guo et al. [30] introduced CAMD, a cross-modal attention mechanism that adaptively learns the relevance and importance of each modality. Rina and Asif [31] proposed the AMFB module based on factorized bilinear pooling, which improves the utilization of complex visual features. Furthermore, Qu et al. [32] encoded multimodal content into quantum states and performed feature fusion using a Quantum Convolutional Neural Network (QCNN) and quantum gate operations, showing promise for future research directions.

Social network analysis-based approaches examine how rumors propagate and how their diffusion structures differ from those of normal information. Rumors often spread faster and more widely, which Guo et al. [33] exploited by designing a model that combines propagation patterns with attention mechanisms. User behaviors—such as reposting, commenting, and liking—can serve as strong indicators. Li et al. [34] enhanced their model by incorporating user credibility scores and attention to prioritize reliable content. Zheng et al. [35] built a social graph connecting posts, comments, and users, using cosine similarity to uncover latent propagation relationships. Moreover, rumor diffusion is characterized by strong temporal dynamics. Wei et al. [36] simulated the temporal structure of Twitter conversations and fused neighboring tweet representations to support dynamic rumor detection effectively.

External fact-based approaches aim to validate the truthfulness of claims by referencing external knowledge. Vlachos and Riedel [37] aligned textual assertions with entries in a knowledge graph to identify factual inaccuracies. Sengan et al. [38] leveraged CNNs to extract stance information from post–reply pairs, aiding in fake news detection. The EANN model by Wang et al. [39] focused on learning event-invariant features to improve generalization across different rumor events. Thorne et al. [40] introduced the FEVER dataset and framework, which uses evidence retrieved from the web to fact-check claims, pushing forward research in automated fact verification. Sun et al. [41] proposed KDCN, a model that aligns multimodal (text-image) features with knowledge graph entities to identify semantic anomalies, thereby enhancing the model's interpretive and inferential power.

## 3. Methodology

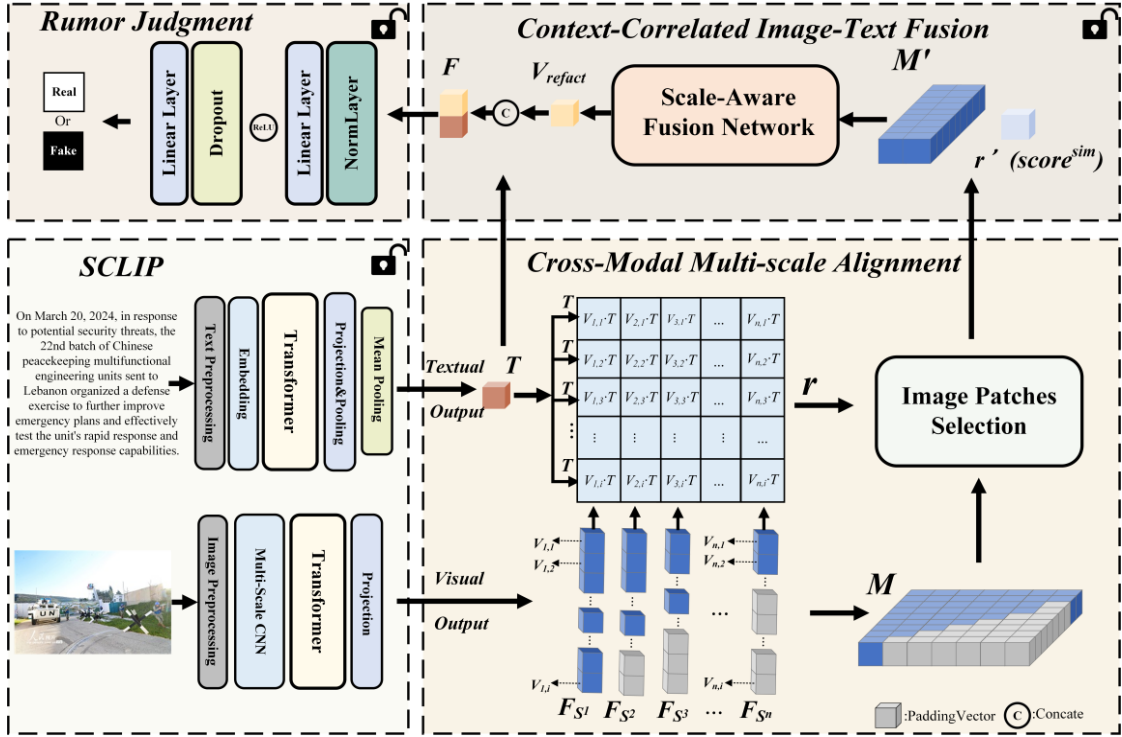

Figure 1. Our proposed MICC consists of four components: Scale-aware Contrastive Language–Image Projection, Cross-Modal Multi-scale Alignment, Context-Correlated Image-Text Fusion, and Rumor Judgment.

### 3.1. Overall Framework

The overall architecture of MICC model is illustrated in Figure 1. This approach identifies image regions that are highly correlated with textual content while also accounting for intra-image variability, thereby improving rumor detection performance.

During inference, given a social media post containing both text and image, the model uses the SCLIP encoder—pretrained via contrastive learning—to encode the text into a global semantic feature $T$ and the image into a set of multi-scale visual embeddings $M$. Based on this, the Cross-Modal Multi-scale Alignment module constructs a dot-product relevance matrix $D$ between the textual feature vector $T$ and the multi-scale visual feature set $M$, and selects the top-K most relevant visual features $M'$ based on their alignment with $T$. Next, the Multi-Scale Patch Attention module learns weights over the features in $M'$ by jointly considering positional and semantic correlations, and performs a weighted summation to produce the refined multi-scale visual representation $V_{refact}$. Finally, the text feature $T$ and $V_{refact}$ are concatenated and passed to the rumor classification module, which outputs a prediction of "true" or "false" via a fully connected layer.

## 3.2. Scale-aware Contrastive Language–Image Projection

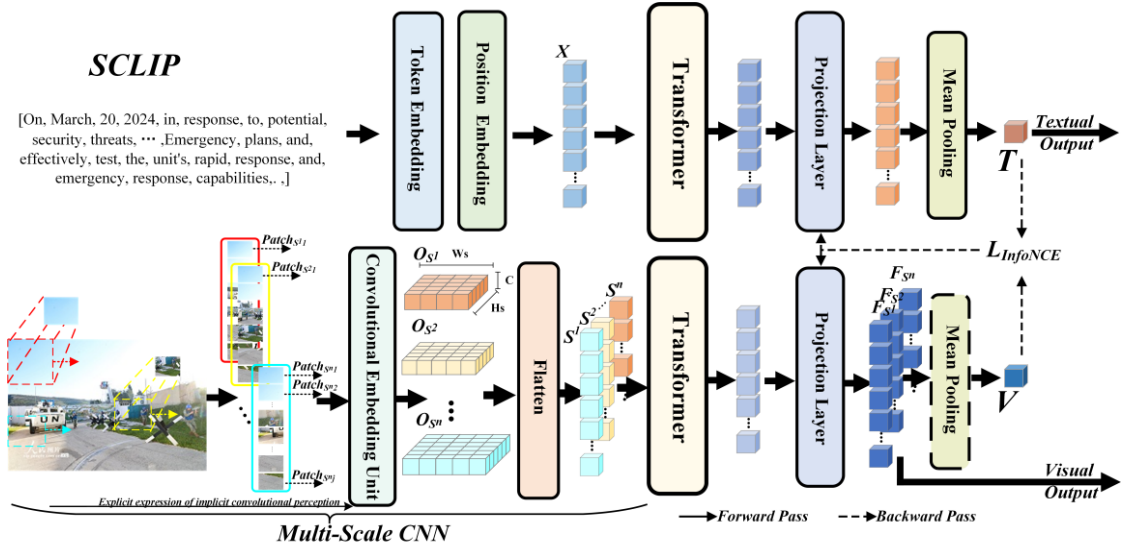

Figure 2. The architecture of SCLIP. The model encodes the input image using multi-scale convolutional receptive fields and extracts deep semantic features through Transformer blocks. Both textual and visual features are projected into a shared semantic space via contrastive learning.

### 3.2.1. Text Embedding

The input text is split into tokens based on the model's vocabulary, ending with a special token, as shown in Equation (1).

$$Text \rightarrow \boldsymbol{TokenSequence}: \{t_1, t_2, \dots, t_n\} \qquad (1)$$

Here, $t_i$ is the token matched to a vocabulary entry.

Each token is converted into a d-dimensional vector via the embedding matrix, represented as $E(t_i)$.

To enable the Transformer to perceive word order, positional encoding is added to supply position information for each sequence element, as shown in Equations (2).

$$\boldsymbol{PE}_{(pos, 2i)} = sin\left(\frac{pos}{10000^{2i/d_{model}}}\right), \boldsymbol{PE}_{(pos, 2i+1)} = cos\left(\frac{pos}{10000^{2i/d_{model}}}\right) \qquad (2)$$

Here, $pos$ is the position, $i$ is the dimension, and $d_{model}$ is the model size. The constant 10000 is used to compute sine and cosine terms so that different positions in the sequence can be distinguished.

Each text token's embedding is added to its position encoding to form the input to the Transformer, as shown in Eq. (3).

$$\boldsymbol{X} = \{X_1, X_2, \dots, X_n\}, \quad X_i = E(t_i) + PE(i) \qquad (3)$$

Based on the Transformer architecture, each token in $X$ is processed through multi-head self-attention mechanisms and feed-forward neural networks, along with residual connections and layer normalization, to capture semantic information from the text.

$$\boldsymbol{H} = Transformer(\boldsymbol{X}) \in R^{n \times d_{model}} \qquad (4)$$

### 3.2.2. Multi-Scale Visual Embedding

To extract semantic representations of images under different receptive fields, a convolutional encoding module called Multi-Scale CNN is designed. This module applies multiple sets of convolutional kernels with varying sizes to the input image to obtain encodings at different spatial scales, followed by semantic learning through a Transformer.

Specifically, given an input image $I \in R^{3 \times W \times H}$ and a set of predefined convolutional receptive fields $\{k_{S^1}, k_{S^2}, ..., k_S\}$, for each receptive field scale $k_{S^i} \times k_{S^i}$, a 2D convolution operation $Conv2D(k_{S^i}, stride = k_{S^i})$ with kernel size and stride both set to $k_{S^i}$ is applied to slide over the image, producing an output feature map:

$$O(S^i) \in R^{C \times H_{S^i} \times W_{S^i}} \tag{5}$$

Here, $H_{S^i} = \frac{H}{k_{S^i}}, W_{S^i} = \frac{W}{k_{S^i}}$

The resulting feature map is then flattened and transformed into a sequence of patch vectors.

$$Q^{(S^i)} = \{q_1^{S^i}, q_2^{S^i}, \cdots, q_j^{S^i}\} = \text{Flatten}(O_{S^i}) \in R^{N_s \times C}, N_s = H_{S^i} \times W_{S^i} \tag{6}$$

Here, each vector $q_j^{S^i} \in R^C$ represents the embedding of a $k_{S^i} \times k_{S^i}$ receptive field in the image. For each receptive field scale $S^i$, a corresponding patch embedding sequence $Q^{(S^i)}$ is generated.

Through this process, the original image is ultimately encoded into multiple sets of convolutional receptive field embeddings, denoted as $S^1, S^2, ..., S^i$. Each embedding vector in $S^i$ corresponds to a local region under a specific receptive field. The set $S = \{S^1, S^2, ..., S^i\}$ represents multi-scale visual features.

The embedding of each region under each convolutional receptive field is used as input to the Transformer to extract features and capture the semantic information of the image.

$$G = \text{Transformer}(S) \in R^{m \times d_{model}} \tag{7}$$

### 3.2.3. Global Semantic Projection with Contrastive Learning

The network introduces a nonlinear projection module after the Transformer output and constructs a cross-modal contrastive learning objective to encourage the model to learn a unified semantic representation across modalities.

As illustrated in the figure, in the text branch, the token sequence encoded by the Transformer $H = \{h_i\}_{i=1}^n \in R^{n \times d_{model}}$ is first passed through a two-layer nonlinear MLP projection head to obtain token-level embeddings in the contrastive space.

$$\tilde{h}_i = \text{MLP}_T(h_i) = W_T^{(2)} \cdot \text{ReLU}\left(W_T^{(1)} \cdot h_i\right) \in R^{d'} \tag{8}$$

Here, $W_T^{(1)}, W_T^{(2)}$ are the weight matrices of the two layers in the text projection module.

Similarly, in the image branch, the image patch embeddings from multi-scale receptive fields

$G = \{q_j^{S^i}\} \in R^{m \times d_{model}}$ are passed through an MLP projection layer with the same structure.

$$\tilde{v}_{i,j} = \text{MLP}_V\left(q_j^{S^i}\right) = W_V^{(2)} \cdot \text{ReLU}\left(W_V^{(1)} \cdot q_j^{S^i}\right) \in R^{d'} \tag{9}$$

Here, $W_V^{(1)}, W_V^{(2)}$ are the weights of the two-layer image projection module.

Then, mean pooling is applied across channels on all projected tokens or patches to derive global semantic features for text and image:

$$T = \frac{1}{n}\sum_{i=1}^{n} \tilde{h}_i, \quad V = \frac{1}{m}\sum_{j=1}^{m} \tilde{v}_{i,j} \tag{10}$$

Here, *n* is the number of text tokens and *m* is the number of image patches.

The network is pretrained using the InfoNCE loss to model similarities between positive and negative image-text pairs.

$$L_{\text{InfoNCE}} = -\frac{1}{N}\sum_{i=1}^{N} \log \frac{\exp(\text{sim}(T_i, V_i)/\tau)}{\sum_{j=1}^{N}\exp(\text{sim}(T_i, V_j)/\tau)} \tag{11}$$

$sim(T, V) = T^{\top} \cdot V$ is the dot-product relevance in the projection space, and $\tau$ adjusts the distinction between positive and negative pairs.

To reduce pretraining costs while retaining the semantic modeling capacity of the original encoders, the network adopts a parameter-freezing strategy during contrastive learning: only the nonlinear projection layers ($\text{MLP}_T, \text{MLP}_V$) for semantic alignment are trained, while the visual encoder (Multi-Scale CNN and image Transformer) and the textual embedding and Transformer modules are frozen.

During the inference and overall model training processes, the network outputs the global textual feature $T$ and the index set of $\tilde{v}_{i,j}$ i.e., the multi-scale visual feature set $M$.

$$M = \{F_{S^1}, F_{S^2}, \ldots, F_{S^i}\}, \quad F_{S^i} = \{v_{i,1}, v_{i,2}, \ldots, v_{i,j}, \ldots\} \tag{12}$$

Where $F_{S^i}$ denotes the feature set corresponding to the i-th convolutional receptive field, and $v_{i,j}$ represents the feature vector of the j-th image patch under receptive field $S^i$. To enable PyTorch batch processing, we apply zero-vector padding to $M$.

During the overall model training process, all module parameters of the network are unfrozen and updated, enabling the encoder to be optimized end-to-end according to the task objective. This allows the model to fully capture deep cross-modal associations and enhance its discriminative capability for downstream tasks.

## 3.3. Cross-Modal Multi-scale Alignment

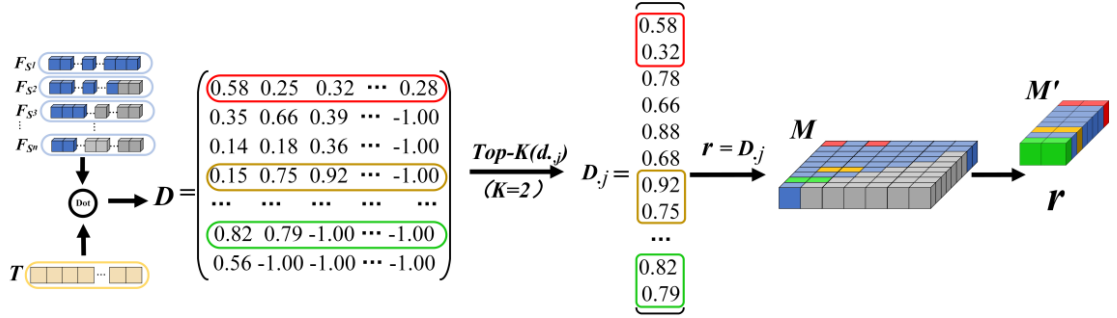

Figure 3. Cross-Modal Image-Text Alignment Module

In multimodal alignment, the goal is to choose image patch $v$ and text $T$ to maximize their mutual information $I(v;T)$, that is, to maximize:

$$I(v; T) = \iint p(v, T) \log\left(\frac{p(v,T)}{p(v)p(T)}\right) dv \, dT \tag{14}$$

However, the joint distribution $p(v,T)$ is generally intractable and cannot be directly computed.

Let $v_a \sim p(v \mid T)$ denote a *positive sample* and $v_b \sim p(v)$ a *negative sample*. The model is expected to distinguish between $p(v \mid T)$ and $p(v)$. Under this formulation, a tractable lower bound of the mutual information can be expressed as:

$$I(v; T) \geq E\left[\log \frac{f(v_a, T)}{\sum_{j=1}^{N} f(v_b, T)}\right] \tag{15}$$

Here, E denotes the expectation, representing the averaging over the scoring outcomes of all text–image pairs to estimate the lower bound of mutual information over the entire data distribution. $f(v,T)$ is a scoring function. If we let $f(v,T) = \exp(v^\top T)$, meaning the dot product is used as the match function, then the above becomes the InfoNCE loss:

$$\mathcal{L}_{InfoNCE} = -E\left[\log \frac{\exp(v_a^\top T)}{\sum_j \exp(v_b^\top T)}\right] \tag{16}$$

This matches the loss function (Eq. (12)) used in SCLIP encoder pretraining. So, in this model, the dot product value can be seen as an increasing estimate of mutual information.

In the implementation of relevance measurement within this module, let the text semantic vector be $T \in R^{d'}$, and for the multi-scale image region set $M = \{v_{i,j}\}$, the feature vector of the j-th image patch at the i-th scale is $v_{i,j} \in R^{d'}$. The relevance between them and the text is computed via dot product as:

$$dot_{i,j} = T^\top \cdot v_{i,j} \tag{14}$$

The above dot product characterizes the directional and magnitude consistency between the text and image patches in the shared semantic space. We compute the relevance between the text and all image patches across all scales and construct a relevance matrix:

$$D = \begin{bmatrix} dot_{1,1} & \cdots & dot_{1,j_1} \\ dot_{2,1} & \cdots & dot_{2,j_2} \\ \vdots & \ddots & \vdots \\ dot_{s,1} & \cdots & dot_{s,j_s} \end{bmatrix} \tag{15}$$

Here, the i-th row of the matrix represents the relevance scores between image patches at scale $S^i$ and the textual feature.

The objective of the alignment task is to maximize the mutual information between all image patches $v$ and the text representation $T$, that is,

$$\max \sum_i I(v_i; T) \tag{16}$$

However, according to the Information Bottleneck Principle, we should not only maximize the mutual information *I(v;T)* between image patch *v* and text *T*, but also compress task-irrelevant redundant information to ensure the compactness and discriminability of the representation. The optimization objective of the information bottleneck can be formulated as:

$$\min I(v; \text{Image}) \quad \text{s.t.} \quad I(v; T) \geq \epsilon \tag{17}$$

Here, *Image* denotes the full set of image regions from the original input, and $\mathbf{v} \subset \mathbf{X}$ represents the subset of selected image patches. The objective is to identify the "minimal yet most informative" semantic region set *M'* that minimizes redundant information while preserving semantic consistency with the text. On this basis, we design a region selection strategy based on dot-product scoring.

We perform a Top-K selection operation on each row of the constructed image-text relevance matrix $D \in R^{S \times J}$, where each row corresponds to all image patches under a specific scale.

$$\text{Top-K}_i = \arg \text{Top-K}_j (\text{dot}_{i,j}) \tag{18}$$

Here, $\text{dot}_{i,j}$ denotes the semantic relevance between the j-th image patch under the i-th scale and the textual feature. The Top-K operation selects the K image regions with the highest scores at each scale.

Under a probabilistic interpretation, according to the InfoNCE theory, when the scoring function is defined as $f(v, T) = \exp(v^\top T)$

$$p(v \mid T) \propto \exp(v^\top T) \tag{19}$$

Specifically, a larger dot product implies a higher conditional posterior probability of a visual region. Thus, selecting the Top-K dot product regions is equivalent to identifying the subset with the highest semantic relevance, which approximates mutual information maximization through a hard selection mechanism, as expressed by the following equivalence:

$$\text{Top-K}(v_{i,j}) \approx \arg\max_{v_{i,j}} p(v_{i,j} \mid T) \approx \arg\max_{v_{i,j}} I(v_{i,j}; T) \tag{20}$$

Finally, all selected region features are aggregated into a visual subset $M' \subset M$, and their corresponding dot product relevance scores form the vector *r*, which serves as key visual representations for subsequent fusion and classification. This mechanism theoretically satisfies the principle of Minimal Sufficient Representation and practically enables effective suppression of noisy regions, thereby improving the accuracy and robustness of the model in multimodal alignment and rumor detection tasks.

## 3.4. Context-Correlated Image-Text Fusion

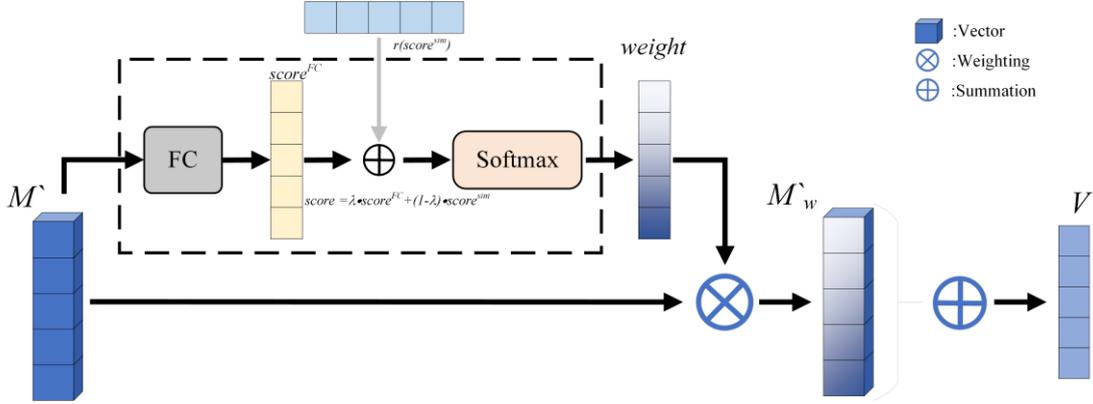

Figure 4. Scale-Aware Fusion network

The Scale-Aware Fusion Network uses a feedforward neural network and a Softmax layer. Each patch is flattened, passed through two linear layers with ReLU activation, and then mapped to an attention score that indicates its importance.

In the two-layer feedforward network, let $w_1 \in R^{h \times d_p}$ be the weight matrix for the i-th patch in $M'$. The attention score is calculated as shown in Equation (23).

$$\text{score}_i^{FC} = \mathbf{w_2^T} \text{ReLU}(\mathbf{w_1} p_i + b_1) + b_2 \tag{19}$$

Where $\mathbf{w_1} \in R^{h \times d_p}$ and $b_1 \in R^h$ are the weight matrix and bias vector from the input layer to the hidden layer, while $\mathbf{w_2} \in R^h$ and $b_2$ denote the weights and bias from the hidden layer to the output layer. $h$ represents the dimensionality of the hidden layer.

The score for each image patch feature vector is computed to form a score vector $\text{score}^{FC}$ which is then combined with the previously obtained relevance score vector $\text{score}^{dot}$ to integrate semantic salience and relevance levels

$$\mathbf{score} = \lambda \mathbf{score^{FC}} + (1 - \lambda) \mathbf{score^{dot}} \tag{20}$$

A Softmax function is further applied to normalize the score vector, ensuring that the sum of weights for all image patch feature vectors equals one. This step is critical for subsequent feature fusion, as it enables the model to dynamically adjust the contribution of each image patch based on its importance. The computation of weight $\alpha_i$ is defined in Equation (24).

$$\alpha_i = \frac{exp(score_i)}{\sum_{j=1}^{n} exp(score_j)} \tag{24}$$

Here, $n$ denotes the total number of image patch feature vectors.

After obtaining the weights, the module performs a weighted summation over all image patch feature vectors to generate the final aggregated visual representation. This process effectively integrates information from all patch vectors while emphasizing segments that are more relevant to the task. The aggregated visual representation $V_{refact}$ is computed as in Equation (25).

$$\mathbf{V_{refact}} = \sum_{i=1}^{n} \alpha_i p_i \tag{25}$$

After obtaining the textual semantic representation $T$ and the aggregated visual representation $V_{refact}$, the two modalities are concatenated to form a fused multimodal representation $F$ as shown in Equation (26). This fusion ensures the integrity of both modalities and prevents the omission of critical information.

$$\mathbf{F} = \mathbf{T} \oplus \mathbf{V}_{\text{refact}} \tag{26}$$

Here, $F$ denotes the fused feature vector obtained via concatenation, and $\oplus$ represents the concatenation operation

## 3.5. Rumor Judgment

To perform the final rumor classification, we design a rumor judgment module based on a fully connected neural network. The input feature $F \in R^d$ is the multimodal fused representation. This module first applies a linear transformation to the input, followed by Dropout during training to randomly deactivate a subset of neurons, thereby mitigating overfitting. The formulation is given in Equation (27):

$$\mathbf{D}_1 = \text{Dropout}(\mathbf{w}_1^T \mathbf{F} + \mathbf{b}_1) \tag{27}$$

Here, $\mathbf{w}_1 \in R^{d \times h}$ denotes the weight matrix, is the bias vector, and $h$ represents the hidden layer dimension.

Subsequently, the module applies a ReLU activation function to the output features to enhance nonlinear modeling capability, followed by a second linear layer for further feature transformation.

$$\mathbf{D}_2 = \mathbf{w}_2^T(\text{RELU}(\mathbf{D}_1)) + \mathbf{b}_2 \tag{28}$$

Here, $\mathbf{w}_2 \in R^{h \times 1}$ denotes the weight matrix and $\mathbf{b}_2 \in R$ the bias vector.

To enhance training stability, a normalization layer is applied to the output features of the second linear layer.

$$\widehat{\mathbf{D}} = \text{NormLayer}(\mathbf{D}_2) \tag{29}$$

Then, the model employs a Sigmoid activation function to normalize the output into a rumor prediction probability, as defined in Equation (30).

$$\hat{y} = \frac{1}{1 + e^{-\widehat{D}}} \tag{30}$$

Here, $\hat{y}$ denotes the rumor probability score output by the model, indicating the likelihood that the input sample is classified as a rumor.

During inference, the final classification result of the model is obtained by applying a threshold to the output probability.

$$\hat{y}_{\text{class}} = \begin{cases} 0, & \text{if } \hat{y} \geq 0.5 \\ 1, & \text{otherwise} \end{cases} \tag{31}$$

To train the entire model, we adopt a binary cross-entropy loss function for end-to-end optimization, aiming to achieve deep multimodal fusion and enhance rumor discrimination capability. The model's output prediction probability is denoted as $\hat{y} \in [0,1]$, representing the likelihood of the input being classified as a rumor, while the ground-truth label is $y \in \{0,1\}$. The cross-entropy loss is computed as shown in Equation (32).

$$\mathcal{L} = -(y \cdot \log(\hat{y}) + (1-y) \cdot \log(1-\hat{y})) \tag{32}$$

Finally, the gradients of all modules in the model are updated through the backpropagation algorithm, thereby enabling joint optimization of multimodal features and accurate prediction for rumor detection. During this stage, we adopt the Adam optimizer to simultaneously update the parameters of all modules.

# 4. Experiments

This chapter details the datasets, experimental setup, and training process used in our study, along with an in-depth analysis of how hyperparameters at various stages affect model performance. We compare the proposed MICC framework against representative state-of-the-art multimodal rumor detection methods, providing strong empirical evidence of its advantages. Ablation experiments are also performed to evaluate the individual contributions of each module in the model.

## 4.1. Experimental Setting

### 4.1.1. Datasets

This study utilizes different datasets in two stages: one for cross-modal contrastive pretraining of the projection layer, and the other for training and evaluating the rumor detection task.

• **Image–Text Pair Datasets:** In the pretraining stage, we use the Flickr30K dataset for English and COCO-CN for Chinese contrastive learning. Table 1 summarizes the dataset statistics. Flickr30K, created by Young et al. [42], consists of 31,783 images, each paired with five English sentence descriptions. COCO-CN [43] is a Chinese image–text alignment dataset based on MS-COCO, containing 20,342 images and roughly 50,000 Chinese captions.

TABLE 1 The statistics of Image–Text Pair datasets.

| Statistic | Images | Text Descriptions | Avg. Descriptions per Image | Language |
|---|---|---|---|---|
| Flickr30K | 31,783 | 158,915 | 5 | English |
| COCO-CN | 20,342 | 50,000 | 2.5 | Chinese |

•**Rumor Detection Datasets:** To evaluate the performance of our model, we adopt two widely used real-world multimodal rumor detection datasets: the Weibo dataset (Song et al. [44]) and the PHEME dataset (Zubiaga et al. [45]). The Weibo dataset is collected from China's popular social media platform Weibo, while the PHEME dataset is composed of English tweets from Twitter. Both datasets include textual content, associated images, and fact-checking annotations. Table 2 provides detailed statistics for both datasets, where "Non-rumors" refers to non-rumorous posts and " Rumors" indicates confirmed rumors. We divide each dataset into training, validation, and testing sets in a ratio of 8:1:1 for model training, tuning, and evaluation.

TABLE 2 The statistics of Rumor Detection datasets.

| Dataset | Non-rumors | Rumors | Images | Language |
|---|---|---|---|---|
| PHEME | 1428 | 590 | 2018 | English |
| WEIBO | 877 | 590 | 1467 | Chinese |

### 4.1.2 Taining Pocess

The training process of this study consists of two stages: (1) cross-modal projection layer pretraining corresponding to each language, and (2) end-to-end training of the full model for the multimodal rumor detection task.

**Stage One: Cross-modal Projection Pretraining (SCLIP Pretraining):**

To enable effective image–text alignment within a unified semantic space, we pretrain the projection module (SCLIP) on English and Chinese image–text datasets separately. The English projection head is trained on the Flickr30K dataset, where the text and image encoders are followed by their respective MLPs and optimized using the InfoNCE loss. Similarly, the Chinese projection head is trained on the COCO-CN dataset, using the same procedure but with Chinese text inputs.

In this stage, only the two projection heads (text MLP and image MLP) are updated, while the rest of the network—including the text Transformer, visual convolution layers, and visual Transformer—remains frozen. The model is trained using the InfoNCE loss, with the Adam optimizer, an initial learning rate of 5e-4, and a batch size of 64.

**Stage Two: End-to-End Training for Multimodal Rumor Detection:**

After completing language-specific projection pretraining, the pretrained modules are used to initialize the full model, which is then trained separately for English and Chinese rumor detection tasks. The English model is trained on the PHEME dataset, initialized with the projection layer pretrained on Flickr30K, and all components—including text and image encoders, alignment module, fusion module, and classification head—are unfrozen for end-to-end optimization. Similarly, the Chinese model is trained on the Weibo dataset, using the projection layer pretrained on COCO-CN and fully fine-tuning all modules.

During training, the model takes paired text and image inputs and outputs a rumor probability, optimized using a binary cross-entropy loss. The Adam optimizer is used with an initial learning rate of 2e-4, and the dataset is split into training, validation, and testing sets with a ratio of 8:1:1.

The goal of this stage is to enhance the model's semantic perception and classification capability in real-world multimodal rumor detection scenarios, ensuring both effective cross-modal alignment and accurate downstream task performance.

Furthermore, we adopt Accuracy, Precision, Recall, and F1-score as the evaluation metrics for the rumor detection task. The formulas for these metrics are provided in Equations (30)–(33).

$$Accuracy = \frac{TP + TN}{TP + TN + FP + FN} \tag{30}$$

$$Precision = \frac{TP}{TP + FP} \tag{31}$$

$$Recall = \frac{TP}{TP + FN} \tag{32}$$

$$F1\ Score = 2 \times \frac{Precision \times Recall}{Precision + Recall} \tag{33}$$

TP (True Positives): The count of instances that are correctly classified as belonging to the positive class. TN (True Negatives): The count of instances that are correctly classified as belonging to the negative class. FP (False Positives): The count of negative instances that are mistakenly classified as positive. FN (False Negatives): The count of positive instances that are mistakenly classified as negative.

## 4.2. Parameter Analysis

This section provides a detailed discussion and experimental validation of the impact of key hyperparameters on model performance at various stages. Specifically, we examine the effect of convolutional receptive fields in SCLIP—including their number and size—on performance; the influence of the Top-K selection value K in Cross-Modal Image-Text Alignment; and in the Scale-Aware Fusion Network, the number of linear layers in the feed-forward network, the dimensionality of their hidden layers, and the effect of the hyperparameter λ. Based on baseline values and definitions listed in Table 3, we conduct ablation studies on each hyperparameter individually to analyze their impact and to identify the optimal configuration.

TABLE 3. Baseline hyperparameter configuration used in parameter analysis. Each hyperparameter is varied independently during ablation studies to evaluate its impact on model performance.

| Stage | Hyperparameter | Default Value | Description |
|---|---|---|---|
| SCLIP | Receptive field sizes | 2 | Kernel sizes used for multi-scale convolution |
| | Number of receptive fields | [32,64] | Number of scales in visual feature extraction |
| Alignment | Top-K visual regions | 2 | Number of image patches selected for alignment |
| Scale-ware Fusion Network | Number of linear layers | 2 | Number of feedforward layers in fusion network |
| | Hidden dimension per layer | 256 | Hidden size for each linear transformation |
| | Score fusion weight | 0.7 | Controls balance between $score^{fc}$ and $score^{sim}$ |

To investigate the impact of the number and scale combinations of convolutional receptive fields in SCLIP on model performance, we conducted systematic experiments on both the WEIBO and PHEME datasets. In the Multi-Scale CNN encoding module, we varied the number of receptive fields and the kernel sizes—selected from a pool of 24, 32, 48, and 64—and compared the resulting performance in terms of Accuracy, Precision, Recall, and F1-score under each configuration.

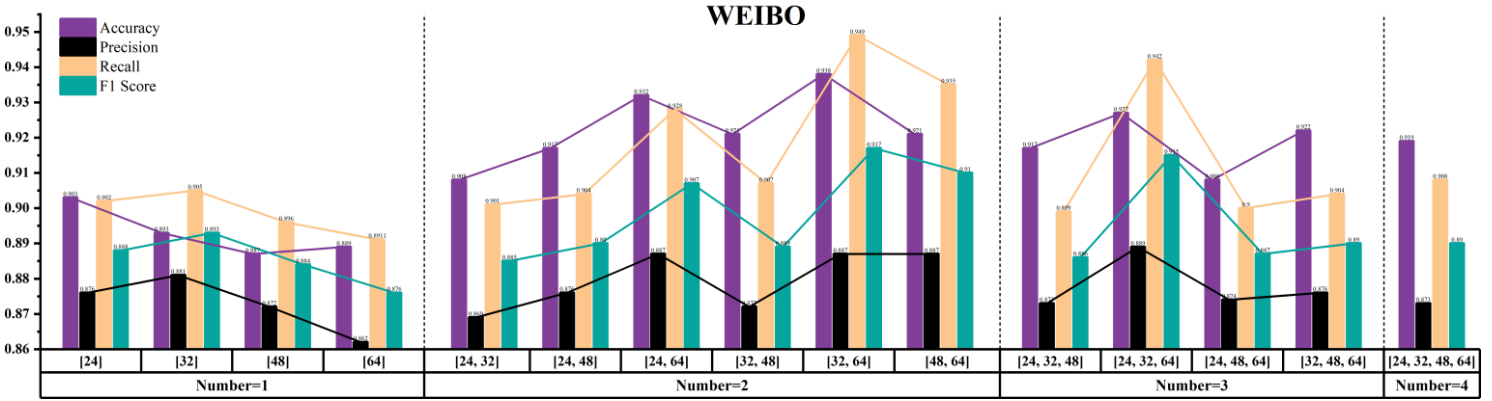

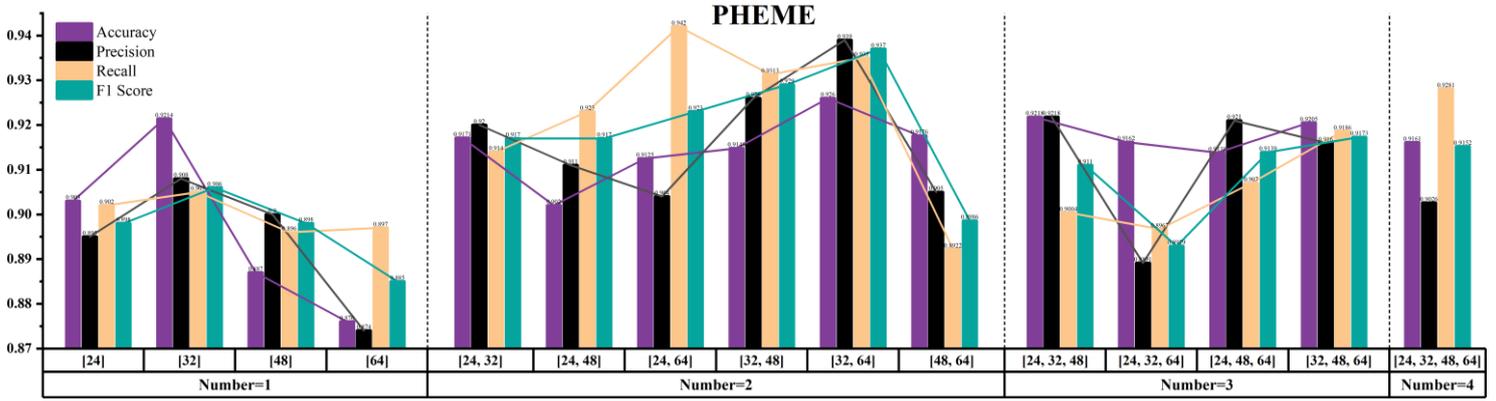

Figure 5. The impact of different numbers and sizes ond sizes of convolutional receptive fields on model performance on the WEIBO and PHEME datasets. The number of receptive fields increases from 1 to 4. The X-axis represents combinations of receptive field sizes, and the Y-axis shows four evaluation metrics: Accuracy, Precision, Recall, and F1 Score.

As illustrated in Figure 5, both the scale and number of receptive fields exert a notable influence on the model's semantic representation and cross-modal alignment capabilities. On the WEIBO dataset, configurations using single-scale receptive fields [24] and [32] achieved better performance than those using [48] and [64], with F1 scores surpassing 0.89, showing the significance of fine-grained features in short-form Chinese social media content. Conversely, the [64] setting alone resulted in reduced performance, indicating that overly coarse visual features may cause the loss of crucial local details. In dual-scale configurations, both [24,64] and [32,64] offered clear performance gains, with [32,64] achieving the highest accuracy of 0.938 on WEIBO, demonstrating the advantage of integrating local and global semantics. On the PHEME dataset, the [32,64] combination reached an F1 score of 0.937, outperforming all other configurations, which supports the benefit of multi-scale complementarity in English-language rumor detection. However, further increasing the number of receptive fields—e.g., [24,32,64] or [24,32,48,64]—did not yield additional improvements and in some cases led to performance drops. This suggests that introducing too many scales may result in redundancy or noise, which hinders semantic alignment and final classification.

To examine how the number of Top-K aligned image regions (K) affects model performance within the Cross-Modal Image-Text Alignment module, we conduct a series of experiments with K ranging from 1 to 8 on the WEIBO and PHEME datasets. The results are illustrated in Figure 6.

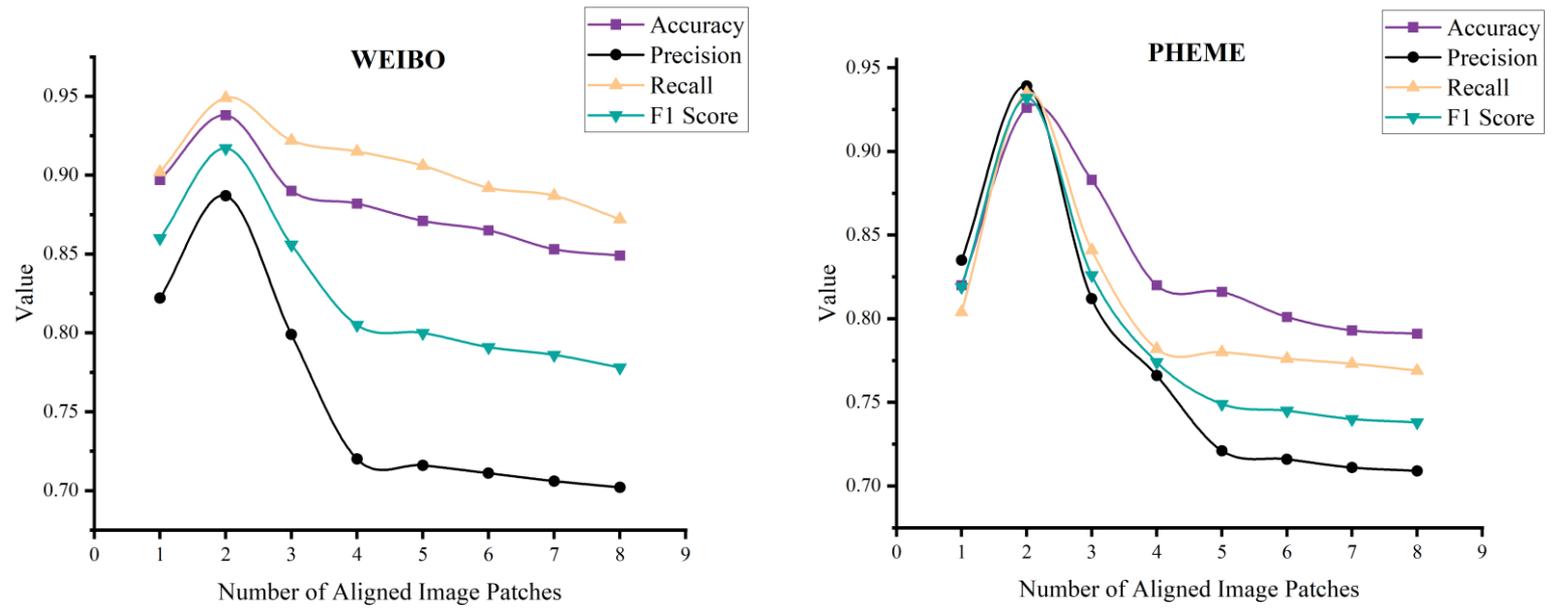

Figure 6. The impact of aligning the number of Top-K regions on rumor detection results

As illustrated in Figure 6, the model performs best in the Cross-Modal Image-Text Alignment module when K is set to 2, corresponding to the selection of the most and second-most relevant image regions. This finding indicates that increasing the number of aligned patches does not always enhance the quality of multimodal representation; on the contrary, it may introduce semantically irrelevant regions that act as noise and undermine the model's detection capability. The result also supports the theoretical foundations of mutual information maximization and the information bottleneck principle.

In the Scale-Aware Fusion Network, we examine how the number of linear layers and the hidden dimension size affect model performance. While deeper networks may capture more complex patterns, excessive depth or width can lead to overfitting. We vary the number of linear layers (2–6) and hidden dimensions (64–2048) in the Multi-Scale Patch Attention feed-forward network and evaluate performance on the WEIBO dataset using Accuracy and F1-score.

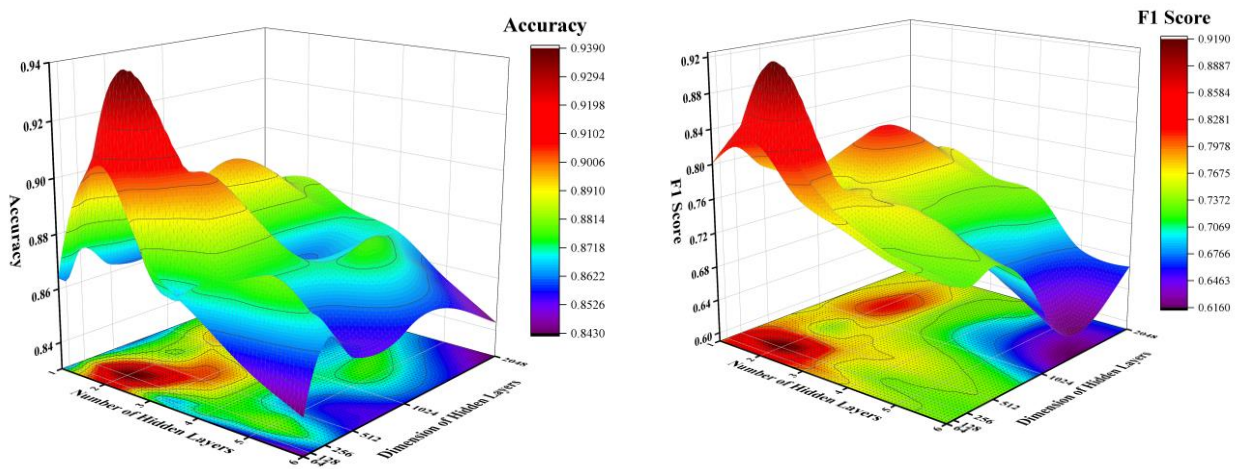

Figure 7. The impact of the number of fully connected layers and the dimension of hidden layers on model performance

Figure 7 shows that model performance improves with increasing linear layers up to 2, but degrades beyond that. The optimal configuration—2 layers and a 256-dimensional hidden size—achieves peak accuracy and F1-score, suggesting a good balance between capacity and generalization. Shallower networks lack semantic depth, while deeper or wider ones (e.g., >3 layers or 2048 dimensions) risk overfitting. Overall, coordinated design of network depth and width is essential to capture key multi-scale visual cues and enhance image–text semantic fusion.

Furthermore, we investigate how the Scale-Aware Fusion Network's hyperparameter λ—which controls the balance between semantic and relevance scores—affects classification performance. Experiments are conducted on both the WEIBO and PHEME datasets, with λ values selected at a step size of 0.05, and Accuracy and F1-score used as evaluation criteria.

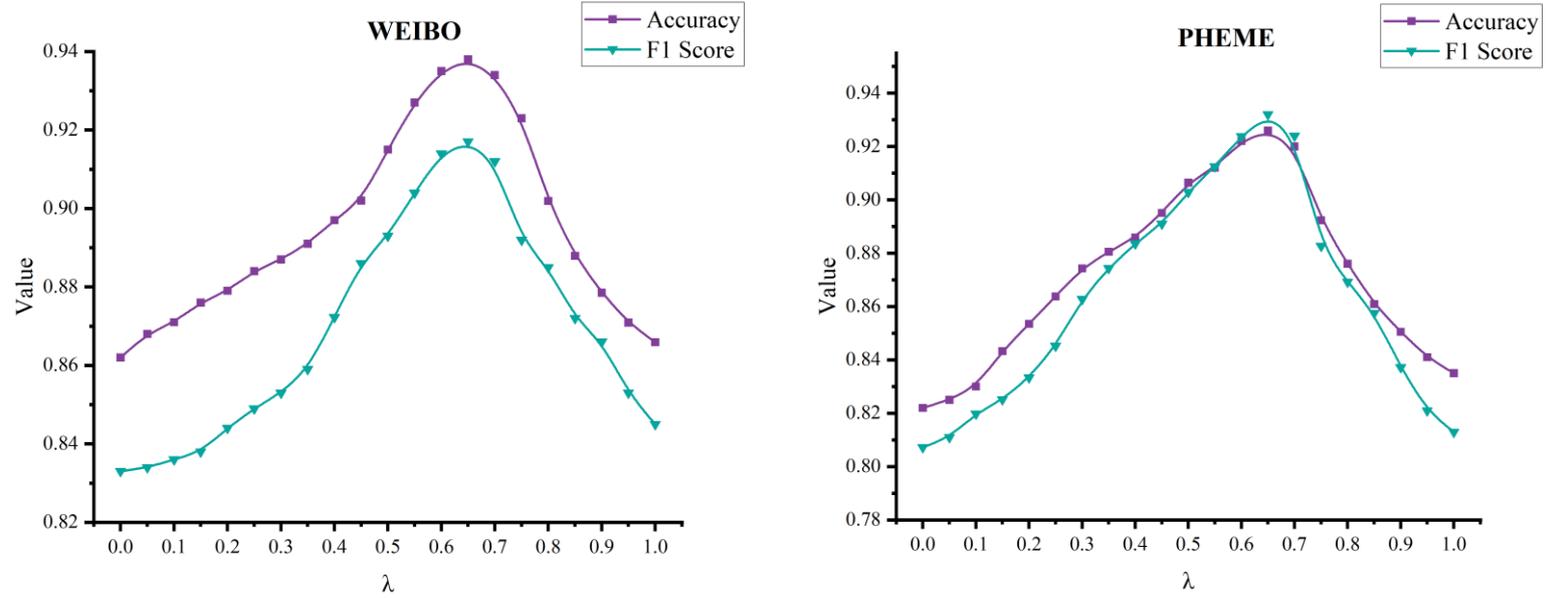

Figure 8. Experimental results of different λ values on WEIBO and PHEME datasets

As shown in Figure 8, the model performs best when λ = 0.65 on Weibo and λ = 0.70 on PHEME. This suggests that emphasizing semantic scores helps highlight key image-text matches and improves detection. However, if λ is too high, irrelevant regions may gain undue attention; if too low, the model may ignore meaningful internal semantics in favor of shallow correlations.

## 4.3. Baselines

To evaluate the effectiveness of MICC, we compare it with several baseline models under both unimodal and multimodal settings.

### 4.3.1. Single Modality Methods

**TextGCN :** TextGCN is a GCN-based approach that models the corpus as a heterogeneous graph of words and documents, learning node representations through graph convolution to capture semantic dependencies for rumor classification [23].

**QSAN:** QSAN is a detection method that integrates quantum-driven text encoding with a novel symbolic attention mechanism, designed specifically for fake news detection [25].

**EBGCN:** EBGCN employs a Bayesian approach to reassess the reliability of latent relationships within the propagation structure, enabling more robust fake news detection [46].

### 4.3.2. Multi Modal Methods

**MVAE:** MVAE employs a variational autoencoder framework to jointly model textual and visual data for fake news detection [28].

**SAFE:** SAFE is a multimodal method that extracts textual and visual features from news content, along with their inter-modal relationships, for fake news detection [47]. It enhances detection by capturing the similarity between modalities.

**MFAN:** MFAN detects rumors by inferring latent links to enhance social graph features and aligning complementary information across text, image, and social graph modalities [35].

**HMCAN:** HMCAN integrates multimodal contextual information and hierarchical textual semantics within a unified deep learning model to detect fake news [8].

**CAF-ODNN:** CAF-ODNN extracts fine-grained cross-modal relationships by introducing image captions, bidirectional complementary attention, and channel normalization, and incorporates an optimized deep neural network for enhanced rumor detection [48].

**MRAN:** MRAN is a multimodal relational attention network that integrates multi-layer textual encodings with VGG19-based visual features, leveraging intra- and inter-modal similarities to generate high-level fusion features for fake news detection [49].

**SePro:** SePro is a supervised graph attention network designed to identify key contextual nodes from propagation and semantic graphs, refining the context before feeding it into a large language model (LLM) for improved rumor detection [50].

## 4.4. Performance Comparison

In the comparative experiments, our model is configured with the optimal hyperparameters as follows: receptive field sizes = 2; receptive field values = [32, 64]; top-K visual regions = 2; number of linear layers = 2; hidden dimension = 256; score fusion weight = 0.65. Table 4 presents the rumor detection results of both baseline models and our proposed model on the WEIBO and PHEME datasets. Based on the experimental outcomes, along with the results in Tables 3 and 4, Analyzing the performance of each compared model with ours, and draw the following conclusions:

TABLE 4 Results of comparison among different models on PHEME and WEIBO datasets.

| | WEIBO | | | | PHEME | | | |
|---|---|---|---|---|---|---|---|---|
| Method | Accuracy | Precision | Recall | F1 | Accuracy | Precision | Recall | F1 |
| TextGCN | 0.789 | 0.843 | 0.779 | 0.809 | 0.828 | 0.801 | 0.782 | 0.791 |
| QSAN | 0.721 | 0.722 | 0.692 | 0.706 | 0.762 | 0.716 | 0.662 | 0.688 |
| EBGCN | 0.836 | 0.862 | 0.822 | 0.842 | 0.834 | 0.825 | 0.799 | 0.812 |
| MVAE | 0.722 | 0.714 | 0.712 | 0.712 | 0.778 | 0.741 | 0.725 | 0.732 |
| SAFE | 0.852 | 0.856 | 0.859 | 0.857 | 0.821 | 0.801 | 0.806 | 0.803 |
| MFAN | 0.889 | 0.890 | 0.881 | 0.885 | 0.887 | 0.881 | 0.862 | 0.871 |
| HMCAN | 0.885 | 0.888 | 0.885 | 0.886 | 0.881 | 0.870 | 0.871 | 0.870 |
| CAF-ODNN | 0.917 | 0.903 | 0.911 | 0.902 | 0.879 | 0.884 | **0.948** | 0.915 |
| MRAN | 0.902 | 0.900 | 0.900 | 0.900 | 0.870 | 0.870 | 0.868 | 0.874 |
| SePro | 0.916 | **0.924** | <u>0.933</u> | **0.928** | <u>0.904</u> | <u>0.897</u> | 0.933 | <u>0.915</u> |
| **Ours** | **0.938** | 0.887 | **0.949** | <u>0.917</u> | **0.926** | **0.939** | <u>0.935</u> | **0.932** |

The MICC model proposed in this paper outperforms all baseline methods on both the WEIBO and PHEME datasets, achieving the highest accuracies of 0.938 and 0.926, and F1 scores of 0.917 and 0.932, respectively. These results significantly exceed those of existing unimodal and multimodal rumor detection models.

From a broader perspective, multimodal methods generally outperform unimodal ones, as the latter lack sufficient modalities to comprehensively represent cross-modal rumor patterns. Classical approaches such as SVM-TS rely heavily on handcrafted features, resulting in poor performance on deep semantic modeling and modality interaction. QSAN, a text-only method, yields F1 scores of merely 0.706 and 0.688, respectively, much lower than multimodal counterparts. Although EBGCN leverages graph structures to improve textual representation and attains F1 scores of 0.842 and 0.812 on the WEIBO and PHEME datasets, it falls short of state-of-the-art multimodal methods due to its exclusion of visual features and susceptibility to graph noise.

Among the bimodal models, MVAE exhibits the weakest performance, with F1 scores of only 0.712 and 0.732 on the WEIBO and PHEME datasets, respectively, indicating that its variational autoencoder-based structure has clear limitations in modeling complex cross-modal semantics. EANN and SAFE achieve F1 scores of 0.857 and 0.885 on WEIBO, and 0.803 and 0.871 on PHEME, demonstrating that the image modality can indeed enhance rumor detection performance, though their image-text alignment or cross-modal interaction mechanisms remain insufficient.

In contrast, MRAN achieves F1 scores of 0.900 and 0.874 on the two datasets, outperforming EANN and SAFE and validating its stronger cross-modal interaction modeling capability. CAF-ODNN reaches a Recall of 0.948 on PHEME, but its Precision is only 0.884, resulting in an F1 score of 0.915—showing outstanding recall but a tendency for misclassification. SePro performs well across both datasets, with F1 scores of 0.928 and 0.915 on WEIBO and PHEME, respectively. However, it still slightly underperforms compared to MICC (F1: 0.917 and 0.932), which achieves a better balance between high recall and precision.

Notably, HMCAN and MFAN, as tri-modal models, achieved F1 scores of 0.886 and 0.885 on the WEIBO dataset, respectively, yet still underperform MICC. This suggests that the inclusion of social information alone is insufficient to compensate for inadequate integration between visual and textual modalities. MICC, despite using only text and image modalities, outperforms all comparison models, including those leveraging three modalities. This advantage is attributed to its cross-modal alignment mechanism and multi-scale visual reconstruction module, which effectively strengthen the deep semantic correlation between text and images and fully exploit the potential of bimodal information.

## 4.5. Ablation Study

In order to assess the contribution of each component in our framework, we construct several simplified versions of our model and compare their performance with relevant baselines. The specific configurations of these variants are described below.

**MICC-MSCM:** Uses the pretrained CLIP model in place of our designed SCLIP extractor and switches to cosine similarity as the alignment reference.

**MICC-A/γ:** Excludes the Image-Text Alignment module, concatenating unaligned image patch vectors and text features directly for classification.

**MICC-A/δ:** Also removes Image-Text Alignment, using concatenated global image and text features instead.

**MICC-Ms/α:** Omits the Scale-Aware Fusion module, directly concatenating aligned image patch features and text features as input.

**MICC-Ms/β:** Similar to Ms/α, but inserts a linear projection to reduce image patch dimensions before concatenation with text.

**MICC-Concate:** Replaces the concatenation-based fusion with the cross-modal attention mechanism from the MFAN framework.

Each variant is tested on key evaluation metrics and benchmarked against the full MICC model. Results are reported in Table 5.

TABLE 5 Experimental results of the variations of MICC.

| Method | | MICC - Ms/α | MICC - Ms/β | MICC - A/γ | MICC - A/δ | MICC - SCLIP | MICC - Concate | MICC |
|---|---|---|---|---|---|---|---|---|
| WEIBO | Acc. | 0.843 | 0.856 | 0.842 | 0.746 | 0.925 | 0.885 | **0.938** |
| | Pre. | 0.759 | 0.769 | 0.763 | 0.775 | 0.848 | 0.836 | **0.887** |
| | Re. | 0.804 | 0.843 | 0.784 | 0.682 | 0.941 | 0.901 | **0.949** |
| | F1. | 0.781 | 0.804 | 0.777 | 0.726 | 0.897 | 0.867 | **0.935** |
| PHEME | Acc. | 0.820 | 0.785 | 0.668 | 0.642 | 0.899 | 0.874 | **0.926** |
| | Pre. | 0.799 | 0.764 | 0.621 | 0.740 | 0.868 | 0.822 | **0.939** |
| | Re. | 0.826 | 0.802 | 0.711 | 0.662 | 0.882 | 0.875 | **0.935** |
| | F1. | 0.812 | 0.814 | 0.663 | 0.699 | 0.875 | 0.848 | **0.932** |

As shown in Table 5, all ablation variants exhibit inferior performance compared to the full MICC model on both the WEIBO and PHEME datasets, confirming the significance of each module in the framework. A comparison between the -Ms and -A groups reveals that the latter consistently underperforms across all evaluation metrics, highlighting the critical role of the image-text alignment module in enhancing cross-modal comprehension. By precisely aligning semantic regions between images and text, this module effectively suppresses redundant or noisy inter-modal information. In the two simplified models -Ms/α and -Ms/β, although -Ms/β introduces linear dimensionality reduction before feature fusion, its performance is slightly inferior to -Ms/α, which directly concatenates features, suggesting that in the absence of explicit fusion mechanisms, textual features contribute more to rumor detection and underscoring the dominant role of the textual modality. Regarding alignment, -A/γ (which uses fine-grained visual features without alignment) outperforms -A/δ (which uses global visual features), demonstrating that fine-grained alignment of multi-scale image regions is more effective than coarse global fusion. Replacing the custom SCLIP encoder with standard CLIP (-SCLIP) also leads to performance degradation, validating the superiority of the multi-scale convolutional structure in SCLIP for fine-grained semantic modeling in rumor detection. Finally, replacing our fusion strategy with MFAN's cross-modal attention in the -Concate variant results in inferior performance across all metrics, indicating that the proposed Scale-Aware Fusion Network better captures semantic dependencies between modalities while suppressing modality inconsistency and redundancy, thereby enhancing the model's discriminative power and robustness.

# 5. Conclusion

To address the limitations of insufficient image-text feature alignment and suboptimal modality fusion in current multimodal rumor detection, this paper proposes a novel cross-modal alignment and fusion framework. Built upon the proposed SCLIP encoder, the method extracts local semantic features from images using multi-scale convolutions with varying receptive fields, and captures global semantic representations of text via a Transformer. Through contrastive learning, both modalities are projected into a shared vector space. A Cross-Modal Multi-Scale Alignment Module is introduced to select semantically relevant visual regions based on the principles of mutual information maximization and the information bottleneck, enabling fine-grained alignment between modalities. Subsequently, a Scale-Aware Fusion Network is constructed to perform weighted fusion based on both visual scale and semantic relevance, enhancing the expressiveness of multimodal representations. Experiments conducted on the Weibo and PHEME datasets demonstrate that the proposed method consistently outperforms state-of-the-art approaches across accuracy, precision, recall, and F1 score, showcasing superior performance and generalization ability in rumor detection tasks. Furthermore, the method offers strong extensibility, as the encoder can be replaced with more powerful multimodal pre-trained models and the fusion mechanism can be adapted to tasks such as sentiment analysis, fake news detection, and visual question answering. Future work will explore lightweight model designs to support real-time deployment on edge devices or mobile platforms.